\documentclass[11pt,a4paper]{article}
\usepackage[hyperref]{emnlp-ijcnlp-2019}
\usepackage{times,latexsym}
\usepackage[shortcuts]{extdash}
\usepackage{url}
\usepackage[T1]{fontenc}
\usepackage{amssymb}
\usepackage{amsmath}
\usepackage{bbm}
\usepackage{breqn}
\usepackage{algorithm}
\usepackage{algpseudocode}
\usepackage{mathptmx}
\usepackage{multicol}
\usepackage{tikz}
\usepackage{booktabs}
\usepackage{tabularx}
\usepackage{dcolumn}
\newcolumntype{d}[1]{D{.}{.}{#1}}
\usetikzlibrary{shapes.geometric}
\usetikzlibrary{arrows.meta}

\DeclareMathAlphabet{\mathcal}{OMS}{cmsy}{m}{n}
\DeclareSymbolFont{matha}{OML}{txmi}{m}{it}
\DeclareMathSymbol{\varv}{\mathord}{matha}{118}

\DeclareMathOperator*{\argmin}{argmin}
\usepackage{url}

\usepackage{xspace,mfirstuc,tabulary}

\aclfinalcopy 

\title{Rotate \emph{King} to get \emph{Queen}: Word Relationships as Orthogonal Transformations in Embedding Space}

\author{Kawin Ethayarajh\thanks{Work partly done at the University of Toronto.} \\
  Stanford University\\
  {\tt kawin@stanford.edu} \\
}

\date{}

\begin{document}
\maketitle
\begin{abstract}
A notable property of word embeddings is that word relationships can exist as linear substructures in the embedding space. For example, \emph{gender} corresponds to $\vec{\textit{woman}} - \vec{\textit{man}}$ and $\vec{\textit{queen}} - \vec{\textit{king}}$. This, in turn, allows word analogies to be solved arithmetically: $\vec{\textit{king}} - \vec{\textit{man}} + \vec{\textit{woman}} \approx \vec{\textit{queen}}$. This property is notable because it suggests that models trained on word embeddings can easily learn such relationships as geometric translations. However, there is no evidence that models \emph{exclusively} represent relationships in this manner. We document an alternative way in which downstream models might learn these relationships: orthogonal and linear transformations. For example, given a translation vector for \emph{gender}, we can find an orthogonal matrix $R$, representing a rotation and reflection, such that $R(\vec{\textit{king}}) \approx \vec{\textit{queen}}$ and $R(\vec{\textit{man}}) \approx \vec{\textit{woman}}$. Analogical reasoning using orthogonal transformations is almost as accurate as using vector arithmetic; using linear transformations is more accurate than both. Our findings suggest that these transformations can be as good a representation of word relationships as translation vectors.

\end{abstract}

\section{Introduction}

Word embeddings are a cornerstone of current methods in NLP. A notable property of these vectors is that word relationships can exist as linear substructures in the embedding space \citep{mikolov2013efficient}. For example, \emph{gender} can be expressed as the translation vectors $\vec{\textit{woman}} - \vec{\textit{man}}$ and $\vec{\textit{queen}} - \vec{\textit{king}}$; similarly, \emph{past tense} can be expressed as $\vec{\textit{thought}} - \vec{\textit{think}}$ and $\vec{\textit{talked}} - \vec{\textit{talk}}$. This, in turn, allows word analogies to be solved arithmetically. For example, \emph{`man is to woman as king is to ?'} can be solved by finding the vector closest to $\vec{\textit{king}} -  \vec{\textit{man}} + \vec{\textit{woman}}$, which should be $\vec{\textit{queen}}$ if one excludes the query words.

\citet{ethayarajh2018towards} proved that when there is no reconstruction error, a word analogy that can be solved arithmetically holds exactly over a set of ordered word pairs iff the co-occurrence shifted PMI is the same for every word pair and across any two word pairs. This means that strict conditions need to be satisfied by the training corpus for a word analogy to hold exactly, and these conditions are not necessarily satisfied by every analogy that makes intuitive sense. For example, most analogies involving countries and their currency cannot be solved arithmetically using Wikipedia-trained skipgram vectors \citep{ethayarajh2018towards}. 

The fact that word relationships \emph{can} exist as linear substructures is still notable, as it suggests that models trained with embeddings can easily learn these relationships as geometric translations. For example, as we noted earlier, it is easy to learn a translation vector $\vec{b}$ such that $\vec{\textit{queen}} = \vec{\textit{king}} + \vec{b}$ and $\vec{\textit{woman}} = \vec{\textit{man}} + \vec{b}$. However, \citeauthor{ethayarajh2018towards}'s proof and corresponding empirical evidence suggest that models do not \emph{exclusively} represent word relationships in this manner. While past work has acknowledged that downstream models can capture relationships as complex non-linear transformations \citep{murdoch2018beyond}, it has not studied whether there are simpler linear alternatives that can also describe word relationships. 

In this paper, we first document one such alternative: orthogonal transformations. More specifically, given the mean translation vector $\vec{b}$ for a word relationship (e.g., \emph{gender}), we can find an orthogonal matrix that represents the relationship just as well as $\vec{b}$. For example, there is an orthogonal matrix $R$ such that $R(\vec{\textit{king}}) \approx \vec{\textit{queen}}$ and $R(\vec{\textit{man}}) \approx \vec{\textit{woman}}$. To find $R$ for a word relationship, we first create a source matrix $X$ of randomly sampled word vectors and a target matrix $Y$ by shifting $X$ by $\vec{b}$. We then use the closed-form solution to orthogonal Procrustes \citep{schonemann1966generalized} to determine $R$, which is the orthogonal matrix that most closely maps $X$ to $Y$. If we broaden our search to include all linear transformations -- not just orthogonal ones -- we can find a matrix $A$ to represent the relationship by using the analytical solution to ordinary least squares.

We find that using orthogonal transformations for analogical reasoning is almost as accurate as using vector arithmetic, and using linear transformations is more accurate than both. However, given that finding the orthogonal matrix analytically is much more computationally expensive, we do not recommend using our method to solve analogies in practice. Rather, our key insight is that there are parsimonious representations of word relationships between the two extremes of simple geometric translations and complex non-linear transformations. Our empirical finding offers novel insight into how downstream NLP models, both large and small, may be inferring word relationships. It suggests that a simple linear regression model or a single attention head of a Transformer \citep{vaswani2017attention} can adequately represent many word relationships, ranging from the one between a country and its capital to the one between an adjective and its superlative form.

\section{Related Work}

\paragraph{Word Embeddings} Word embeddings are distributed representations in a low-dimensional continuous space. They can capture semantic and syntactic properties of words as linear substructures, allowing relationships to be expressed as geometric translations \citep{mikolov2013distributed}. Word vectors can be learned from: (a) neural networks that learn representations by predicting co-occurrences \citep{bengio2003neural,mikolov2013distributed}; (b) low-rank approximations of word-context matrices containing a co-occurrence statistic \citep{landauer1997solution,levy2014neural}. 

\paragraph{Solving Word Analogies} There are two main strategies for solving word analogies, \texttt{3CosAdd} and \texttt{3CosMul} \citep{levy2014linguistic}. The former is the familiar vector arithmetic method: given a word analogy task \emph{a:b::x:?}, the answer is $ \argmin_w \cos(\vec{w}, \vec{x} + \vec{b} - \vec{a})$. For \texttt{3CosMul}, the answer is $\argmin_w (\cos(\vec{w}, \vec{x}) \cos(\vec{w}, \vec{b}))/(\cos(\vec{w}, \vec{a}) + \epsilon)$, where $\epsilon$ prevents null division. Although \texttt{3CosMul} is more accurate on average, we do not discuss it further because it does not create a distinct representation of the relationship. As noted previously, our goal is \emph{not} to come up with a better strategy for solving analogies, but to show that there are parsimonious representations of word relationships other than translation vectors.

\paragraph{Orthogonal Maps} Orthogonal transformations have been applied to word embeddings to achieve various objectives. Most famously, they have been used for the cross-lingual alignment of word embeddings trained on non-parallel data, for unsupservised machine translation \citep{conneau2017word}. \citet{rothe2016ultradense} proposed a method for creating ultra-dense word embeddings in more meaningful subspaces by first learning an orthogonal transformation of the embeddings and then clipping all but the relevant dimensions. \citet{park2017rotated} built on this work, exploring several strategies for rotating embeddings to obtain more semantically meaningful dimensions. However, to our knowledge, orthogonal transformations themselves have not been used to represent word relationships; our work is novel in this respect.

\begin{table*}[t]
    \centering 
    
    \small
    \begin{tabularx}{\textwidth}{Xcccccc}
        \toprule \multicolumn{1}{c}{Analogy Category} & \multicolumn{3}{c}{Accuracy} & \multicolumn{3}{c}{Avg Cosine Similarity with Solution} \\ \cmidrule(lr){2-4} \cmidrule(lr){5-7}
        & Orthogonal & Linear & Translative & Orthogonal & Linear & Translative  \\ \midrule
capital-common-countries & \bf 0.957 & \bf 0.957 & \bf 0.957 & 0.802 & 0.844 & 0.847 \\
capital-world & 0.922 & \bf 0.966 & \bf 0.966 & 0.727 & 0.786 & 0.793 \\
currency & 0.300 & \bf 0.467 & 0.267 & 0.517 & 0.515 & 0.511 \\
city-in-state & 0.529 & 0.897 & \bf 0.926 & 0.705 & 0.775 & 0.802 \\
family & \bf 0.913 & \bf 0.913 & \bf 0.913 & 0.840 & 0.840 & 0.859 \\
gram1-adjective-to-adverb & 0.438 & \bf 0.500 & \bf 0.500 & 0.667 & 0.670 & 0.678 \\
gram2-opposite & \bf 0.621 & 0.586 & 0.517 & 0.629 & 0.607 & 0.632 \\
gram3-comparative & 0.865 & 0.865 & \bf 0.892 & 0.791 & 0.768 & 0.812 \\
gram4-superlative & \bf 0.912 & 0.882 & \bf 0.912 & 0.747 & 0.705 & 0.764 \\
gram5-present-participle & 0.909 & \bf 0.939 & 0.848 & 0.813 & 0.808 & 0.829 \\
gram6-nationality-adjective & 0.902 & 0.902 & \bf 0.927 & 0.816 & 0.837 & 0.841 \\
gram7-past-tense & 0.600 & \bf 0.650 & 0.625 & 0.768 & 0.770 & 0.775 \\
gram8-plural & 0.892 & \bf 0.919 & 0.892 & 0.796 & 0.787 & 0.807 \\
gram9-plural-verbs & \bf 0.900 & 0.733 & 0.800 & 0.786 & 0.773 & 0.803 \\ \midrule
Avg & 0.761 & \bf 0.798 & 0.782 & 0.743 & 0.749 & 0.768 \\ \bottomrule
    \end{tabularx}
    \caption{The accuracy on our word analogy task -- detailed in section 4.1 -- when word relationships are represented as orthogonal, linear, and translative functions. The highest accuracy for each category is in bold. As seen in the last row, on average, orthogonal transformations are almost as accurate as translations (0.761 vs.\ 0.782), and the average cosine similarity between a transformed vector and the solution is about the same for both. }
    \label{tab:results}
\end{table*}

\section{Representing Word Relationships}

To formalize the notion of a word relationship, we treat it as an invertible transformation that can hold over an arbitrary number of ordered pairs, following a similar framing proposed by \citet{ethayarajh2018towards} for word analogies. For example, the word pairs \emph{\{(Berlin, Germany), (Paris, France), (Ottawa, Canada)\}} all express the same word relationship because some function $f$ maps each capital city to its respective country. In this paper, we look at three specific types of transformations: \emph{translative}, \emph{orthogonal}, and \emph{linear}. 

\paragraph{Definition 1} \emph{A word relationship $f$ is an invertible transformation that holds over a set of ordered pairs $S$ iff $\forall\ (x,y) \in S, f(x) = y \wedge f^{-1}(y) = x$.} 

\paragraph{Definition 2.1} \emph{A translative word relationship $f$ is a transformation of the form $\vec{x} \mapsto \vec{x} + \vec{b}$, where $\vec{b}$ is the translation vector. $f$ holds over ordered pairs $S$ iff $\forall\ (x,y) \in S, \vec{x} + \vec{b} = \vec{y}$.}

\paragraph{Definition 2.2} \emph{An orthogonal word relationship $f$ is a transformation of the form $\vec{x} \mapsto R\vec{x}$, where $R^TR = I$. $f$ holds over ordered pairs $S$ iff $\forall\ (x,y) \in S, R\vec{x} = \vec{y}$.}

\paragraph{Definition 2.3} \emph{A linear word relationship $f$ is a transformation of the form $\vec{x} \mapsto A\vec{x}$, where $A$ is a non-degenerate square matrix. $f$ holds over ordered pairs $S$ iff $\forall\ (x,y) \in S, A\vec{x} = \vec{y}$.} 

\paragraph{} Given a set of word pairs $S$, how can we determine a translative, orthogonal, or linear transformation $f$ such that $\forall\ (x,y) \in S, f(\vec{x}) \approx \vec{y}$? Fortunately, there are closed form solutions for each case. To define a translation, we can simply take the mean of the pairwise difference vectors as our translation vector:
\begin{equation}
    \vec{b} = \frac{1}{|S|} \sum_{(x,y) \in S} \vec{y} - \vec{x}
    \label{bias_def}
\end{equation}
For orthogonal transformations, we first uniformly randomly sample $n$ words from the vocabulary and stack their word vectors to get a source matrix $X$. Then, we add $\vec{b}$ to each sampled word vector to get a target matrix $Y$. Finding the orthogonal matrix that most closely maps $X$ to $Y$ is called the orthogonal Procrustes problem:
\begin{equation}
    R = \argmin_{\Omega}\ \| \Omega X - Y \|_F\ \text{s.t. } \Omega^T\Omega = I
    \label{r_def}
\end{equation}
Orthogonal Procrustes has a closed-form solution that we can use to find $R$ \citep{schonemann1966generalized}. We frame the problem similarly to find a linear map $A$ that does not necessarily need to be orthogonal. If we assume that the linear transformation should minimize the ordinary least squares objective,
\begin{equation}
\begin{split}
    A &= \argmin_{\Omega}\ \| \Omega X - Y \|^2 \\
    &= Y X^T (XX^T)^{-1} 
\end{split}
\label{a_def}
\end{equation}
Note that our approach to finding the orthogonal and linear transformations is to find those that best approximate the geometric translation by $\vec{b}$. The reasoning behind this is simple: in practice, the number of word pairs in $S$ is much smaller than the embedding dimensionality $d$, so trying to find a map $x \mapsto y\ \forall\ (x,y) \in S' \subset S$ would be very conducive to over-fitting. Since we randomly sample $n$ words to create $X$ and $Y$, we can choose $n \gg d$ to prevent this problem. 

Although we are ultimately learning to approximate a translation, representing a word relationship as an orthogonal matrix as opposed to a translation vector has some useful mathematical properties, such as preserving the inner product: $\forall\ (x,y) \in S, \left< \vec{x}, \vec{y} \right> = \left< R\vec{x}, R\vec{y} \right>$. \citet{ethayarajh2018towards} proved that when there is no reconstruction error, the word-context matrix $M$ that is implicitly factorized by models such as skipgram and GloVe can be recovered from the inner products of word vectors. Since the inner product is preserved under rotation, $M$ can also be recovered from an orthogonally transformed word space. The same does not hold under translation.

\section{Evaluating the Representations}

\begin{figure*}
    \centering
    \minipage{0.50\textwidth}
      \includegraphics[width=\linewidth]{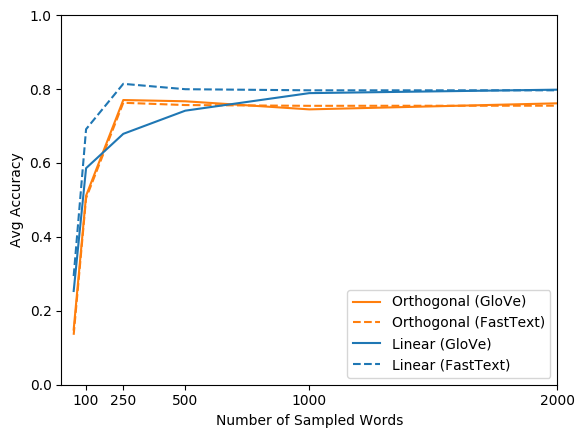}
    \endminipage\hfill
    \minipage{0.50\textwidth}%
      \includegraphics[width=\linewidth]{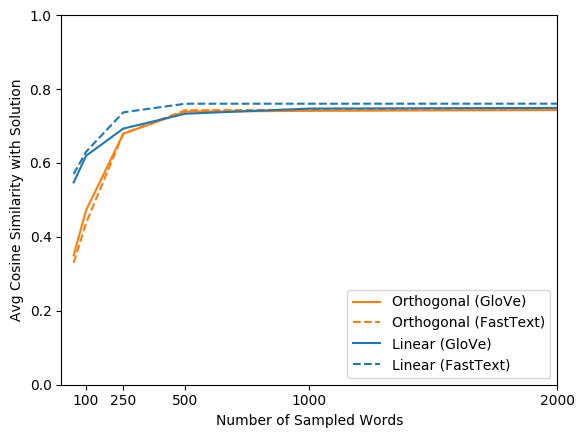}
    \endminipage
    \caption{The accuracy on our word analogy task (left) and the average cosine similarity between the predicted and actual answers (right) as $n$, the number of sampled words used to learn the transformation, increases. The accuracy plateaus for $n \geq 250$ and the similarity plateaus for $n \geq 500$, suggesting that a robust transformation can be learned with relatively little data. Use of GloVe vs.\ FastText vectors makes no difference as $n \to 2000$. }
    \label{fig:figures}
\end{figure*}

\subsection{Task and Setup}

We evaluate the different representations of word relationships using analogy tasks. However, we do not aim to solve word analogies in the traditional sense, since that largely depends on which words are present in each analogy's 4-tuple. Instead, we first calculate the mean translation vector $\vec{b}$ by averaging difference vectors across all word pairs, as defined in (\ref{bias_def}). $\vec{b}$ is also used to estimate matrices for the orthogonal and linear transformations (see (\ref{r_def}) and (\ref{a_def})). We then create a set of word pairs for each analogy category: e.g., \{\emph{(Berlin, Germany), (Paris, France)}, ... \} for \emph{country-capital}. Each type of transformation -- translative, orthogonal, and linear -- is evaluated by how accurately it maps source words to target words in this set of word pairs. We use pre-trained GloVe vectors \citep{pennington2014glove} and $n = 2000$ for our main results in Table \ref{tab:results} and repeat our experiments with FastText vectors \citep{bojanowski2017enriching} in Figure \ref{fig:figures}. We source our analogies from \citet{mikolov2013efficient}, as it contains a diverse set of categories. 

\subsection{Results}

As seen in Table \ref{tab:results}, orthogonal transformations are almost as accurate as geometric translations on our word analogy task: the average accuracy is 0.761 and 0.782 respectively. Linear transformations are more accurate than both (0.798). We would expect linear transformations to outperform orthogonal ones, given that the set of possible linear transformations is a superset of the set of possible orthogonal transformations. However, it is surprising that linear maps also outperform geometric translations, given that they are ultimately learned using translation vectors.

In the right half of Table \ref{tab:results}, we list the average cosine similarity between the transformed source vector and the actual target vector (e.g, $\cos(R(\vec{\textit{king}}), \vec{\textit{queen}})$). This is to mitigate concerns that because the transformed source vector is mapped to the closest word vector, orthogonal transformations are only accurate due to the sparsity of the word space. As seen in Table \ref{tab:results}, such concerns would be unfounded: the average cosine similarity for orthogonal transformations across all categories is 0.743, almost as high as the 0.768 for translations. This suggests that even if we considered a larger portion of the vocabulary as candidate answers, or if the word space were denser, orthogonal transformations would still be almost as accurate as translations on our task.

Our only hyperparameter is $n$, the number of randomly sampled words used to generate $X$ and $Y$. As seen in Figure \ref{fig:figures}, the accuracy plateaus for $n \geq 250$ and the average cosine similarity with the target vector plateaus for $n \geq 500$. This suggests that it is possible to learn orthogonal and linear transformations representing word relationships with relatively little data. As $n \to 2000$, differences in performance between GloVe and FastText disappear, though linear transformations are more accurate than orthogonal ones for all $n$. This also highlights why the translation vector $\vec{b}$ is used to learn the transformations instead of a subset of the actual word pairs: for most analogy categories, there are fewer than 250 pairs in the dataset, and learning with so few word pairs would lead to poor accuracy.

\subsection{Implications}

\paragraph{Evaluating Embeddings} The literature has often evaluated the quality of word embeddings by testing their ability to solve word analogies arithmetically \citep{mikolov2013distributed,pennington2014glove}. If word relationships were \emph{exclusively} geometric translations, this would be reasonable. However, given that word relationships can also be orthogonal or linear transformations, the usefulness of these tests as a measure of embedding quality should be reconsidered. Other arguments, both theoretical and empirical, have been made in the past against the use of analogies for evaluation  \citep{schluter2018word,drozd2016word,rogers2017too}.

\paragraph{Model Architecture} Given that bias terms are not necessarily needed to learn word relationships, the architecture of downstream models trained on word embeddings can be modified accordingly. Transformers \citep{vaswani2017attention} already make extensive use of linear maps in multi-headed attention and appear to be justified in doing so. Moreover, recent work has found that certain attention heads are sensitive to certain syntax, positional information, and other linguistic phenomena \cite{voita2019analyzing,clark2019does}. For  example, \citet{clark2019does} identified heads  that attend to the direct objects of verbs, noun determiners, and coreference mentions with surprisingly high accuracy. However, these studies have not examined whether semantic word relationships -- such as gender -- also correspond to certain attention heads. Given that our findings suggest that individual attention heads have the capacity to learn such relationships, this is a promising direction for future work.

\paragraph{Bias in Word Embeddings} The most common method for removing gender bias in word embeddings involves defining a bias subspace in the embedding space and then subtracting from each word vector its projection on this subspace \citep{bolukbasi2016man}. Under certain conditions, this method can provably debias skipgram and GloVe word embeddings \citep{ethayarajh2019understanding}, but in practice, these conditions are typically not satisfied and gender associations can still be recovered from the embedding space \cite{gonen2019lipstick}. Our findings in this paper suggest another way in which downstream models may be learning such biases, by representing gender as an orthogonal or linear transformation. This, in turn, may help explain the existence of such bias in contextualized word representations as well \citep{zhao2019gender}. Given that these transformations are another way in which social biases can manifest, exploring more diverse strategies for debiasing -- or alternatively, understanding the limits of debiasing strategies -- is another direction for future work.

\section{Conclusion}

Word relationships in embedding space are generally thought of as simple geometric translations or complex non-linear transformations. However, we found that there are parsimonious representations of relationships between these two extremes, namely orthogonal and linear transformations. In addition, we found that it is possible to easily learn an orthogonal or linear transformation for a word relationship given its mean translation vector. Analogical reasoning done using linear transformations is in fact more accurate than using geometric translations. This finding offers novel insight into how downstream NLP models may be inferring word relationships. For example, it suggests that a single attention head in a Transformer has sufficient capacity to represent a semantic or syntactical word relationship, concurring with recent findings that certain attention heads have syntax- and position-specific behavior.

\section*{Acknowledgments}

We thank the anonymous reviewers for their insightful comments. We thank the Natural Sciences and Engineering Research Council of Canada (NSERC) for their financial support.

\bibliography{naaclhlt2019}
\bibliographystyle{acl_natbib}
\clearpage

\appendix

\end{document}